\documentclass[letterpaper]{article} 
\usepackage[preprint]{aaai2027}  
\usepackage[hyphens]{url}  
\usepackage{graphicx} 
\usepackage{pdfpages}
\usepackage[table]{xcolor}
\usepackage{multirow}
\usepackage{natbib}  
\usepackage{caption} 
\usepackage{amsmath}
\usepackage{amssymb}
\usepackage{booktabs}
\usepackage{microtype}
\usepackage{array}
\usepackage{tabularx}
\usepackage{algorithm}
\usepackage{algpseudocode}
\usepackage{placeins} 
\usepackage{dblfloatfix}
\usepackage{needspace}
\usepackage{cuted}

\newif\ifcolorRQ
\colorRQtrue
\DeclareRobustCommand{\rqhighlight}[2]{%
  \ifcolorRQ
    \begingroup
      \setlength{\fboxsep}{1.5pt}%
      \colorbox{#1}{\strut #2}%
    \endgroup
  \else
    #2%
  \fi
}
\DeclareRobustCommand{\rqinlinehighlight}[2]{%
  \ifcolorRQ
    \begingroup
      \setlength{\fboxsep}{1pt}%
      \smash{\colorbox{#1}{\strut #2}}%
    \endgroup
  \else
    #2%
  \fi
}

\newcommand{\AppHardOOD}{Hard/OOD Capacity Results}
\newcommand{\AppMechanism}{Mechanism Ablations}
\newcommand{\AppMTU}{MTU-Bench Transfer Results}
\newcommand{\AppFAIL}{FAIL-TaLMs Readiness and Failure Handling}
\newcommand{\AppBFCL}{Call-Realization Boundary: BFCL}
\newcommand{\AppEfficiency}{Efficiency Trade-off}

\newcommand{\AppSliceConstruction}{Development Slice Construction}

\newcommand{\appsecref}[1]{the supplement section \emph{#1}}
\newcommand{\RQTypedState}{RQ5}

\microtypesetup{protrusion=true,expansion=true}
\DeclareCaptionFont{aaairoman}{\normalfont\normalsize}
\renewcommand{\arraystretch}{1.04}
\definecolor{tableband}{RGB}{232,241,249}
\newcolumntype{Y}{>{\centering\arraybackslash}X}
\title{From State to Action: OODA-Tool
for Reliable Multi-Turn Tool Use}

\author{
Rongfeng Guo\textsuperscript{\rm 1}\equalcontrib,
Yinxuan Huang\textsuperscript{\rm 2}\equalcontrib,
Yusen Wu\textsuperscript{\rm 3},
Maoqing Zhong\textsuperscript{\rm 4},
\\
Yunlu Chen\textsuperscript{\rm 5},
Meng Tang\textsuperscript{\rm 6},
Teng Long\textsuperscript{\rm 7},
Vincent Tao Hu\textsuperscript{\rm 1}
}
\affiliations{
\textsuperscript{\rm 1}Huazhong University of Science and Technology\\
\textsuperscript{\rm 2}Knowin AI, Shenzhen, China\\
\textsuperscript{\rm 3}Fujian University of Technology\\
\textsuperscript{\rm 4}Jiangxi University of Science and Technology\\
\textsuperscript{\rm 5}King Abdullah University of Science and Technology\\
\textsuperscript{\rm 6}University of California, Merced\\
\textsuperscript{\rm 7}University of Amsterdam
}

\begin{document}

\maketitle

\begin{abstract}
Reliable multi-turn tool use requires an agent to preserve an evolving
task state and ensure that each action remains consistent with it.
However, direct function-calling and ReAct-style policies learn state
tracking and action generation within the same autoregressive
trajectory. This coupling creates \textbf{state-action competition}:
the pressure to produce the next call can overwrite or ignore
information accumulated earlier in the interaction. Inspired by Boyd's Observe--Orient--Decide--Act cycle, we introduce \textbf{OODA-Tool}, a typed closed-loop policy designed to mitigate this
competition by separating state preservation from action realization.
Rather than generating an action directly from the interaction history,
OODA-Tool routes each decision through controller-checked intermediate
states, ensuring that the final output remains grounded in the current
task state. Specifically, \emph{Observe} reconstructs the task state,
\emph{Orient} determines whether execution is warranted,
\emph{Decide} forms an admissible action structure, and \emph{Act} realizes the
external output. We evaluate OODA-Tool against direct function-calling and ReAct
policies using Qwen3 models ranging from 0.6B to 14B across multi-turn,
multi-tool, and incomplete-information settings. OODA-Tool consistently improves task success across model sizes, with larger gains on smaller models and on tasks whose actions depend
strongly on information accumulated across turns and prior tool
results. Controlled variants, stage-level ablations, and transfer
evaluations further demonstrate the robustness of these improvements.
\end{abstract}

\section{Introduction}

Language-model agents increasingly rely on search engines, databases,
code executors, and business APIs to solve tasks beyond parametric
knowledge
\citep{DBLP:journals/corr/abs-2112-09332,
DBLP:conf/nips/SchickDDRLHZCS23,
DBLP:journals/csur/QinHLCDCZZHXHFSWQTZLSXZ25,
DBLP:conf/emnlp/LiZ000YLHL23,
DBLP:conf/nips/YangPNY23,
DBLP:conf/iclr/0036YZXLL0DMYZ024,
DBLP:conf/iclr/ZhouX0ZLSCOBF0N24}.
Multi-turn tool use is difficult because the validity of a call depends
on an interaction state that changes as the task unfolds.
Evaluations show that single-turn performance may not transfer to
stateful, long-horizon interaction
\citep{DBLP:conf/iclr/00020LCYPJ24,
DBLP:conf/acl/0003H0XLZHLPW25,
DBLP:journals/corr/abs-2406-12045,
DBLP:conf/acl/AcikgozGDYZEKHT25}.
Syntactic
validity and eventual task completion are therefore incomplete
indicators of correct behavior
\citep{DBLP:conf/icml/PatilMYJSSG25}. Step-wise evaluations reveal
failures hidden by aggregate outcomes
\citep{DBLP:conf/acl/ChenDZLLZZZLCZ24}.
A call may be well formed yet rely
on stale or unsupported information. Likewise, final success can mask
inconsistent intermediate decisions, creating a gap between outcome
correctness and consistency with the evolving state
\citep{DBLP:conf/naacl/LuHZANBMMLYWP25,
DBLP:conf/emnlp/OuGGNY25}.

\begin{figure}[!t]
    \centering
    \includegraphics[width=0.92\columnwidth]
    {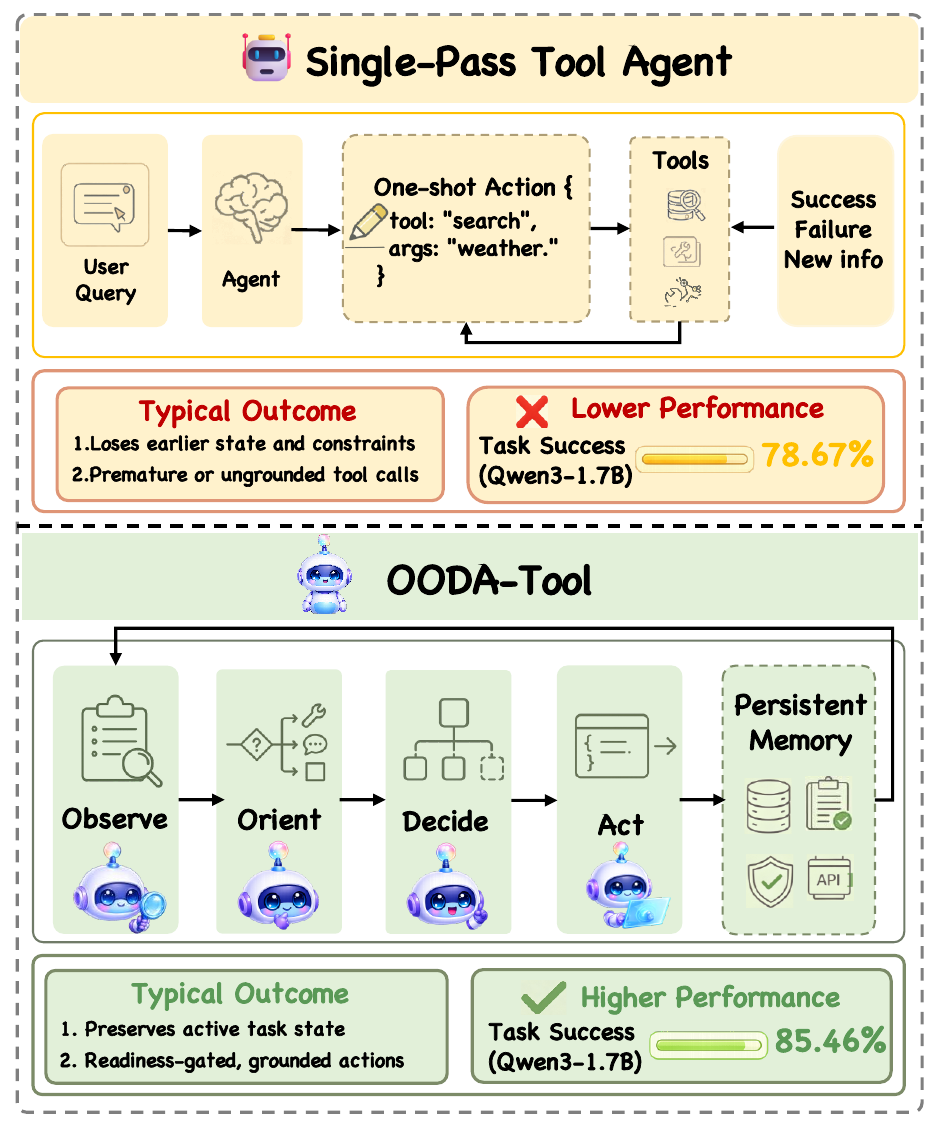}
    \caption{Comparison between a single-pass tool agent and our staged OODA-Tool agent.}
    \label{fig:intro_motivation}
    \vspace{-5pt}
\end{figure}

Yet most agent designs leave the connection between accumulated state
and the next action implicit
\citep{DBLP:conf/nips/PatilZ0G24,
DBLP:conf/iclr/QinLYZYLLCTQZHT24}.
StateFlow externalizes part of this control through explicit
state-machine transitions \citep{DBLP:journals/corr/abs-2403-11322,DBLP:journals/corr/abs-2602-02486}.
As shown in Figure~\ref{fig:intro_motivation}, direct
function-calling policies map the interaction history to a concrete
call within one generation. ReAct-style policies make the reasoning
process visible, but represent it as free-form reasoning rather than as
controller-checkable state
\citep{DBLP:conf/iclr/YaoZYDSN023,DBLP:conf/nips/SchickDDRLHZCS23, DBLP:conf/icml/ErdoganL0MFAKG25}.
In both cases, the model may respond appropriately to the latest turn
while failing to carry an earlier constraint or state update into the
next action. We call this failure mode \emph{state-action competition}:
the immediate demand to produce an action interferes with preserving
and applying the task state accumulated across turns.

A useful lens on this failure comes from Boyd's
Observe--Orient--Decide--Act (OODA) cycle in military command and
control \citep{boyd2018discourse,DBLP:conf/acl/BaiZHWLWYGC26}. OODA treats decision making as a
closed loop rather than a direct reaction to the latest observation.
New information is first interpreted against the current operational
picture, after which an action is selected and executed. The
consequences of that action then update the operational picture for the
next cycle. Its relevance to tool use lies not simply in dividing
reasoning into four steps, but in maintaining a coherent state between
observation and action. Figure~\ref{fig:intro_motivation} illustrates
this contrast: the upper pipeline leaves this coordination inside a
monolithic generation, whereas the lower pipeline makes the decision
state explicit and returns tool feedback to the next cycle
\citep{DBLP:conf/nips/ShinnCGNY23, DBLP:journals/corr/abs-2510-02837,
DBLP:conf/nips/MaZZYYJLKH24}.

Following this principle, we propose \textbf{OODA-Tool}, a typed policy
with explicit supervision at each stage. \emph{Observe} constructs a
provenance-aware representation of the current task, and \emph{Orient}
determines whether the available state supports execution. Based on
this judgment, \emph{Decide} specifies an admissible action structure,
which \emph{Act} realizes as a schema-valid tool call or an authorized
user-facing response. As depicted in the lower half of
Figure~\ref{fig:intro_motivation}, a central controller checks each
handoff before allowing the action to proceed and incorporates the
result into the next cycle. This differs from $\alpha$-UMi, which
primarily divides planning, calling, and summarization across separate
models \citep{DBLP:conf/emnlp/ShenLCYQCZ024}. Plan-and-Act similarly
separates planning from execution. OODA-Tool instead uses
typed interfaces to regulate how accumulated state is converted into
action~\citep{DBLP:conf/iclr/QinLYZYLLCTQZHT24}. This design also raises a central question: whether compressing
history into typed states loses information, or whether stronger
interface constraints suppress otherwise valid actions. We investigate
whether state preservation and action realizability can instead be
improved jointly, as examined in \RQTypedState.

This mechanism yields a clear empirical prediction. OODA-Tool should
provide the greatest benefit when success depends on carrying state
across turns or coordinating sequential tool dependencies. The benefit
should diminish as backbone capacity increases and remain modest when
the main difficulty lies in expanding many parallel calls. The results
follow this pattern. Across Qwen3 models with 0.6B, 1.7B, 4B, 8B, and 14B
parameters, Specialized OODA improves Task Success over
Direct-LoRA by 6.86, 6.79, 6.99, 5.94, and 4.48 points, respectively. The gains are larger
on the hard and out-of-distribution slices, but smaller on simple or
highly parallel calls. Direct-LoRA remains the cheaper choice when
one-pass latency is the primary concern.
Our contributions are as follows:
\begin{itemize}
    \item We propose OODA-Tool, a typed four-stage policy that separates state reconstruction, execution gating, action planning, and grounded realization.
    
    \item We design controlled joint, shared, and specialized variants, together with stage ablations and typed-state bottleneck analyses, to disentangle the effects of typed supervision, multi-pass inference, and stage specialization, and to examine how state preservation supports downstream action realization.
    
    \item We conduct a cross-scale evaluation on ToolDial and three additional benchmarks, demonstrating consistent gains on state-intensive tasks while identifying parallel-call realization as a key limitation.
\end{itemize}
\section{Related Work}
\label{sec:related_work}

\paragraph{Tool learning and function calling.}
Tool learning covers instruction understanding, decomposition, selection,
call generation, and response synthesis
\citep{DBLP:journals/csur/QinHLCDCZZHXHFSWQTZLSXZ25}. Representative
approaches include self-supervised call insertion in Toolformer
\citep{DBLP:conf/nips/SchickDDRLHZCS23}, executable or simulated API-use data
in API-Bank and ToolAlpaca
\citep{DBLP:conf/emnlp/LiZ000YLHL23,DBLP:journals/corr/abs-2306-05301}, and
large-catalog retrieval and call generation in Gorilla and ToolLLM
\citep{DBLP:conf/nips/PatilZ0G24,DBLP:conf/iclr/QinLYZYLLCTQZHT24}.
ToolACE, ToolFlow, and ToolDial increase call complexity and multi-turn
coherence through verified or synthetic interactions
\citep{DBLP:conf/iclr/Liu0ZHYL0GLY0WN25,DBLP:conf/naacl/WangZLLWSJLW25,DBLP:conf/iclr/ShimSLJ25},
while ReAct exposes free-form interleaved reasoning and actions
\citep{DBLP:conf/iclr/YaoZYDSN023}.

\paragraph{Stateful interaction and clarification.}
ToolSandbox introduces stateful execution, implicit dependencies,
conversation, and insufficient-information tasks
\citep{DBLP:conf/naacl/LuHZANBMMLYWP25}; $\tau$-bench scores the resulting
database state under domain policies
\citep{DBLP:journals/corr/abs-2406-12045}. MTU-Bench spans single- and
multi-turn, single- and multi-tool settings, whereas FAIL-TaLMs isolates
underspecification and unavailable-tool failures
\citep{DBLP:conf/iclr/WangWWLSPDZWPZG25,DBLP:conf/naacl/TrevinoCNNW25}.

\paragraph{Agent tuning, modular planning, and constrained execution.}
FireAct, AgentTuning, and Agent-FLAN train agent capabilities from trajectories
or capability-aware data
\citep{DBLP:journals/corr/abs-2310-05915,DBLP:conf/acl/ZengLLWLD024,DBLP:conf/acl/ChenLWZLLCZ24}.
Chain-of-thought, self-consistency, Tree of Thoughts, and Reflexion expose,
aggregate, search, or revise intermediate reasoning
\citep{DBLP:conf/nips/Wei0SBIXCLZ22,DBLP:conf/iclr/0002WSLCNCZ23,DBLP:conf/nips/YaoYZS00N23,DBLP:conf/nips/ShinnCGNY23},
but generally do not define typed execution contracts. The closest modular
baseline separates planning, calling, and summarization
\citep{DBLP:conf/emnlp/ShenLCYQCZ024}; earlier systems connect language models
to external modules or specialized-model controllers
\citep{DBLP:journals/corr/abs-2205-00445,DBLP:conf/nips/0001ST00Z23}.
LLMCompiler schedules function dependencies, and ToolDec constrains syntax
\citep{DBLP:conf/icml/KimMTLMKG24,zhang2024dontfinetunedecodesyntax}.

\paragraph{Evaluation scope.}
BFCL evaluates simple, multiple, parallel, and stateful function calling
\citep{DBLP:conf/icml/PatilMYJSSG25}, while AgentBench and WebArena cover
broader interactive settings
\citep{DBLP:conf/iclr/0036YZXLL0DMYZ024,DBLP:conf/iclr/ZhouX0ZLSCOBF0N24}.
Unlike these approaches, OODA-Tool supervises accumulated evidence,
execution readiness, action dependencies, and argument realization through
typed checkpoints before execution.

\section{Methodology}
This section presents OODA-Tool, including its typed decision process,
stage supervision, and state-sufficiency analysis.

\subsection{Problem Formulation}

\paragraph{Task and interaction history.}
We study a language-model agent that interacts with an external tool
environment over multiple turns. At turn $t$, the agent observes the
ordered interaction history
\begin{equation}
    h_t = \big\langle
    (u_1,a_1,o_1),\ldots,(u_{t-1},a_{t-1},o_{t-1}),u_t
    \big\rangle,
\end{equation}
where $u_i$ is a user message, $a_i$ is an agent tool call or
natural-language response, and $o_i$ is the subsequent environment
observation, which may be empty when no tool is invoked. Let
$\mathcal{T}$ denote the available tool library, including tool names,
documentation, and argument schemas. A monolithic policy samples the
next external output directly:
\begin{equation}
    a_t \sim \pi_{\theta}(\cdot \mid h_t,\mathcal{T}),
\end{equation}
thereby coupling task-state reconstruction, execution-readiness
assessment, action-structure selection, and schema-grounded realization
inside one implicit representation.

\begin{figure*}[t!]
\centering
\includegraphics[width=0.92\textwidth]
{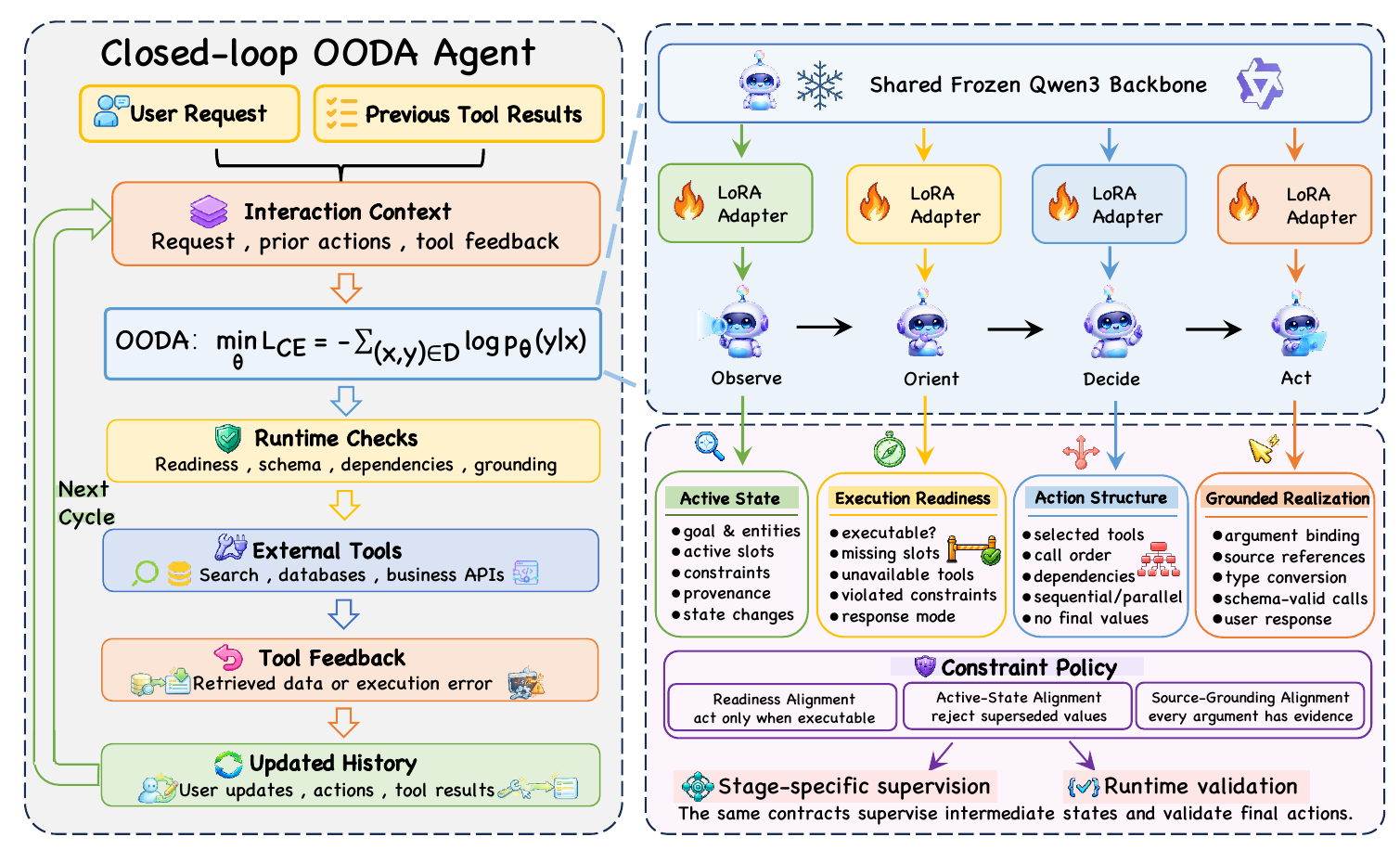}
\caption{Overview of OODA-Tool. The left panel shows the closed-loop
tool-use process under central-controller validation. The upper-right
panel illustrates Specialized OODA with a separate LoRA adapter for
each stage over a shared frozen Qwen3 backbone. The lower-right panel
summarizes the typed intermediate states and the unified constraint
policy used for stage supervision and final-action validation.}
\label{fig:ooda_architecture}
\end{figure*}

\paragraph{Typed OODA policy.}
The typed interfaces and their central-controller checks are summarized
in Figure~\ref{fig:ooda_architecture}. OODA-Tool represents the decision
process as three typed intermediate states followed by a realized
action. In the default history-visible setting,
\begin{alignat}{2}
    &z_t^O &&= f_O(h_t,\mathcal{T}), \\
    &z_t^R &&= f_R(z_t^O,h_t,\mathcal{T}), \\
    &z_t^D &&= f_D(z_t^O,z_t^R,h_t,\mathcal{T}), \\
    &a_t   &&= f_A(z_t^O,z_t^R,z_t^D,h_t,\mathcal{T}),
\end{alignat}
where $z_t^O$, $z_t^R$, and $z_t^D$ are the Observe state,
readiness state, and action structure, respectively. Let
$\mathcal{Z}_D(z_t^R)$ denote the set of action structures permitted by
the readiness state. The central controller enforces
\begin{equation}
    z_t^D \in \mathcal{Z}_D(z_t^R),
\end{equation}
so that Decide cannot authorize a tool structure when Orient requires
clarification, recovery, direct response, or termination. This
decomposition changes neither the environment nor the visible
information; it changes how the decision is represented, supervised,
and checked before the external output is emitted.

\subsection{The OODA-Tool Framework}

At each turn, OODA-Tool executes Observe, Orient, Decide, and Act in
sequence under central-controller validation. A tool result, execution
error, or user update is then appended to the interaction history and
initiates the next cycle.

\paragraph{Observe: task-state reconstruction.}
Observe converts the history into a provenance-aware task state
with fixed field semantics. It records the current goal, entities,
active values, evidence sources, constraints, unfinished subgoals,
recent tool facts, and state changes. Only information available before the current action is included in $h_t$;
$g_t$ provides current-turn supervision but excludes post-action observations.

\paragraph{Orient: execution readiness.}
A correct task state does not necessarily imply that the agent should
act. Orient predicts one of five response modes:
\texttt{SOLVABLE\_WITH\_TOOL},
\texttt{NEED\_CLARIFICATION},
\texttt{RESPOND\_DIRECTLY},
\texttt{RECOVER\_FROM\_FAILURE}, or
\texttt{DONE}. It also records missing slots, violated constraints,
and unavailable tools. Orient determines whether action-structure
selection is admissible and blocks execution when the available
evidence is insufficient.

\paragraph{Decide: action structure.}
Conditioned on the readiness state, Decide instantiates the permitted
action structure. For executable turns, it selects a single call, a
sequential chain, or a parallel action set and specifies the target
tools, call order, and data dependencies without expanding final
argument values. For non-tool modes, it preserves the authorized
clarification, direct-response, recovery, or termination branch for
Act. This separation keeps execution-readiness and action-structure
errors distinct from argument-realization errors.

\paragraph{Act: schema-grounded realization.}
Act compiles the selected action structure into executable calls. It
binds values from the active task state or completed tool returns,
performs schema-permitted deterministic transformations, resolves
references to earlier outputs, and verifies required keys and formats.
In non-tool modes, it emits the authorized clarification question,
recovery action, direct answer, or termination response. Parallel and
repeated calls place the greatest burden on this stage because one
action structure must be expanded into multiple correctly bound call
objects.

\begin{algorithm}[t]
\caption{Specialized OODA-Tool Training and Inference}
\label{alg:ooda_tool}
\small
\begin{algorithmic}[1]

\Require Trajectories $\mathcal{D}$, tool library $\mathcal{T}$,
constraint profile $c$
\Ensure Trained parameters
$\Theta_c^*=\{\theta_{s,c}^*\}_{s\in\{O,R,D,A\}}$

\Statex \textbf{Offline Training}
\State Initialize $\mathcal{D}_{s,c}\gets\emptyset$
for all $s\in\{O,R,D,A\}$

\ForAll{trajectory $\xi\in\mathcal{D}$}
    \ForAll{turn $t$ in $\xi$}
        \State $(h_t,g_t)\gets\Call{CurrentTurnView}{\xi,t}$
        \State $z_t^{O*}\gets\Call{ObserveTarget}{h_t,c}$
        \State $z_t^{R*}\gets\Call{OrientTarget}
        {z_t^{O*},\mathcal{T},g_t,c}$
        \State $z_t^{D*}\gets\Call{DecideTarget}
        {z_t^{O*},z_t^{R*},g_t,c}$
        \State $a_t^*\gets\Call{ActTarget}
        {z_t^{O*},z_t^{R*},z_t^{D*},g_t,c}$
        \State Add the four stage examples to
        $\{\mathcal{D}_{s,c}\}_{s\in\{O,R,D,A\}}$
    \EndFor
\EndFor

\ForAll{$s\in\{O,R,D,A\}$}
    \State $\theta_{s,c}^*
    \gets\Call{SupervisedTrain}{\mathcal{D}_{s,c}}$
\EndFor

\Statex \textbf{Online Inference}
\State Initialize $h_1\gets\langle u_1\rangle$

\For{$t=1,\ldots,H$}
    \State $z_t^O\gets f_O(h_t,\mathcal{T};\theta_{O,c}^*)$
    \State $z_t^R\gets f_R(z_t^O,h_t,\mathcal{T};\theta_{R,c}^*)$
    \State $z_t^D\gets f_D(z_t^O,z_t^R,h_t,\mathcal{T};
    \theta_{D,c}^*)$
    \State Validate $z_t^D\in\mathcal{Z}_D(z_t^R)$
    \State $a_t\gets f_A(z_t^O,z_t^R,z_t^D,h_t,\mathcal{T};
    \theta_{A,c}^*)$
    \State Validate the schema and grounding of $a_t$

    \If{$z_t^R=\texttt{SOLVABLE\_WITH\_TOOL}$}
        \State $o_t\gets\mathcal{E}(a_t)$
        \State $h_{t+1}\gets h_t\oplus(a_t,o_t)$

    \ElsIf{$z_t^R=\texttt{RECOVER\_FROM\_FAILURE}$}
        \State $o_t\gets\mathcal{E}(a_t)$
        \State $h_{t+1}\gets h_t\oplus(a_t,o_t)$

    \ElsIf{$z_t^R=\texttt{NEED\_CLARIFICATION}$}
        \State Emit $a_t$ and receive $u_{t+1}$
        \State $h_{t+1}\gets h_t\oplus(a_t,u_{t+1})$

    \Else
        \State \Return $a_t$
        \Comment{\texttt{RESPOND\_DIRECTLY} or \texttt{DONE}}
    \EndIf
\EndFor

\end{algorithmic}
\end{algorithm}

\subsection{Stage Parameterization and Supervision}
\label{sec:role_supervision}

We instantiate the typed interfaces as \textbf{Joint OODA}, which
serializes the Observe, Orient, Decide, and Act outputs in one model
call; \textbf{Shared OODA}, which uses four stage-wise calls with one
shared LoRA adapter; and \textbf{Specialized OODA}, which uses a
separate LoRA adapter for each stage over the same frozen backbone
\citep{DBLP:conf/iclr/HuSWALWWC22}. These variants isolate the effects
of typed supervision, stage-wise inference, and stage-specific
parameterization.

Typed target construction is controlled by a
\textsc{ConstraintPolicy}. For a profile $c$, the policy specifies
which execution-readiness conditions, active-task-state relations, and
admissible argument sources are represented in the targets. For each
trajectory $\xi$ and turn $t$, we construct the visible prefix $h_t$
and a current-turn gold annotation $g_t$. The annotation contains the
gold response mode, action structure, and realized output for turn
$t$, but excludes future user turns, the observation produced by the
current gold action, and facts revealed only by that observation.
Targets are derived in stage order:
\begin{align}
    z_t^{O*}
    &= \operatorname{ObserveTarget}(h_t,c), \\
    z_t^{R*}
    &= \operatorname{OrientTarget}
       (z_t^{O*},\mathcal{T},g_t,c), \\
    z_t^{D*}
    &= \operatorname{DecideTarget}
       (z_t^{O*},z_t^{R*},g_t,c), \\
    a_t^*
    &= \operatorname{ActTarget}
       (z_t^{O*},z_t^{R*},z_t^{D*},g_t,c).
\end{align}
This construction prevents upstream targets from depending on
information obtained only after the gold action is executed.

For Joint OODA, the supervised target is the serialized sequence
\begin{equation}
    y_t^* =
    \big[z_t^{O*};z_t^{R*};z_t^{D*};a_t^*\big],
\end{equation}
\begingroup
\setlength{\abovedisplayskip}{0pt}
\setlength{\belowdisplayskip}{0pt}
\setlength{\abovedisplayshortskip}{0pt}
\setlength{\belowdisplayshortskip}{0pt}
and training minimizes
\begin{equation}
    \mathcal{L}_{\mathrm{joint}}
    = -\sum_{(x,y)\in\mathcal{D}_{\mathrm{joint},c}}
    \log p_{\theta_c}(y\mid x).
\end{equation}
For Shared and Specialized OODA, stage-level datasets
$\mathcal{D}_{s,c}$ are constructed for
$s\in\{O,R,D,A\}$. Training minimizes
\begin{equation}
    \mathcal{L}_{\mathrm{multi}}
    = -\sum_{s\in\{O,R,D,A\}}
      \sum_{(x,z)\in\mathcal{D}_{s,c}}
      \log p_{\theta_{\rho(s),c}}(z\mid x),
\end{equation}
\endgroup
where $\rho(s)=\mathrm{shared}$ for Shared OODA and
$\rho(s)=s$ for Specialized OODA. The backbone, token-level
cross-entropy objective, and target-construction pipeline remain fixed
across constraint profiles; only the typed relations generated by the
selected profile change. At inference time, the matching profile
configures interface validation, contract retries, and
execution-readiness-based rerouting in the central controller.

\paragraph{State sufficiency.}
The default history-visible setting gives every stage access to $h_t$
in addition to typed upstream states. To test whether the typed
states retain the information required downstream, the
typed-state bottleneck removes history access from Orient,
Decide, and Act, leaving only typed upstream states and tool schemas.
A small degradation under this intervention indicates that the Observe
state retains most task-relevant history and that the readiness state
carries actionable information rather than only a post-hoc
explanation.
\section{Experiments}
We ask whether OODA improves tool use across model scales
(\rqinlinehighlight{red!15}{RQ1}), when its gains are largest
(\rqinlinehighlight{yellow!15}{RQ2}), which components drive its gains
(\rqinlinehighlight{green!15}{RQ3}), how well it transfers and at what cost
(\rqinlinehighlight{blue!15}{RQ4}), whether typed states remain actionable
(\rqinlinehighlight{violet!15}{RQ5}), and how typed state supports argument
grounding (\rqinlinehighlight{orange!15}{RQ6}).
\label{sec:experiments}


\newcolumntype{M}{
  >{\hspace*{1pt}}l<{\hspace*{4pt}}
}
\newcolumntype{S}{
  >{\hspace*{3pt}}c<{\hspace*{3pt}}
}
\newcolumntype{N}{
  >{\hspace*{1.25pt}}r<{\hspace*{1.25pt}}
}
\newcolumntype{G}{
  >{\hspace*{5.5pt}}c
}
\newcolumntype{E}{
  >{\hspace*{5pt}}c
}

\begin{table*}[t!]
\centering
\scriptsize
\renewcommand{\arraystretch}{1.03}
\setlength{\tabcolsep}{0pt}

\resizebox{0.985\textwidth}{!}{%
\begin{tabular}{
@{}
M
S
*{5}{N}
G
*{5}{N}
G
*{5}{N}
E
@{}
}

\toprule

\multirow{2}{*}{\textbf{Method}}
&
\multirow{2}{*}{
  \shortstack[c]{
    \textbf{Structure}\\[-0.3ex]
    \textbf{/ adapters}
  }
}
&
\multicolumn{5}{c}{\textbf{Task Success} $\uparrow$}
&
&
\multicolumn{5}{c}{\textbf{Tool Exact} $\uparrow$}
&
&
\multicolumn{5}{c}{\textbf{Ask--Act Accuracy} $\uparrow$}
&
\\

\cmidrule(lr){3-7}
\cmidrule(lr){9-13}
\cmidrule(lr){15-19}

&
&
\textbf{0.6B}
&
\textbf{1.7B}
&
\textbf{4B}
&
\textbf{8B}
&
\textbf{14B}
&
&
\textbf{0.6B}
&
\textbf{1.7B}
&
\textbf{4B}
&
\textbf{8B}
&
\textbf{14B}
&
&
\textbf{0.6B}
&
\textbf{1.7B}
&
\textbf{4B}
&
\textbf{8B}
&
\textbf{14B}
&
\\

\midrule

\textcolor{gray}{Base Instruct}
&
\textcolor{gray}{--}
&
\textcolor{gray}{0.10}
&
\textcolor{gray}{0.43}
&
\textcolor{gray}{3.04}
&
\textcolor{gray}{7.26}
&
\textcolor{gray}{11.63}
&
&
\textcolor{gray}{0.20}
&
\textcolor{gray}{0.67}
&
\textcolor{gray}{6.18}
&
\textcolor{gray}{13.84}
&
\textcolor{gray}{20.45}
&
&
\textcolor{gray}{69.50}
&
\textcolor{gray}{73.76}
&
\textcolor{gray}{78.42}
&
\textcolor{gray}{82.15}
&
\textcolor{gray}{85.27}
&
\\

\midrule

Direct-LoRA
&
1 adapter
&
78.24
&
78.67
&
80.31
&
83.58
&
90.42
&
&
88.36
&
82.41
&
84.29
&
87.31
&
96.21
&
&
94.06
&
95.03
&
96.08
&
97.05
&
98.04
&
\\

ReAct-LoRA
&
1 adapter
&
65.43
&
73.18
&
79.62
&
81.35
&
84.77
&
&
71.56
&
77.21
&
82.39
&
85.44
&
90.83
&
&
91.32
&
94.71
&
95.26
&
96.59
&
97.84
&
\\

Direct-SC@4
&
4 samples
&
78.90
&
78.90
&
81.12
&
83.92
&
90.72
&
&
88.70
&
82.68
&
84.57
&
87.45
&
96.34
&
&
94.10
&
95.07
&
96.15
&
97.12
&
98.09
&
\\

\addlinespace[1.2pt]

$\alpha$-UMi
&
3 roles
&
\underline{82.90}
&
\underline{83.94}
&
\underline{85.80}
&
\underline{88.10}
&
\underline{93.90}
&
&
\underline{97.90}
&
\underline{98.60}
&
\underline{98.72}
&
\underline{98.94}
&
\underline{99.15}
&
&
\underline{98.35}
&
\underline{98.70}
&
\underline{98.80}
&
\underline{98.92}
&
\underline{99.04}
&
\\

\addlinespace[1.2pt]
\midrule
\multicolumn{20}{l}{\textit{Ours}}
\\
\addlinespace[0.8pt]

OODA
&
Joint
&
83.90
&
84.16
&
86.10
&
88.45
&
94.10
&
&
98.10
&
98.20
&
98.55
&
98.85
&
99.15
&
&
97.95
&
98.22
&
98.45
&
98.70
&
98.90
&
\\

OODA
&
Shared
&
84.55
&
84.78
&
86.68
&
88.96
&
94.50
&
&
98.35
&
98.58
&
98.80
&
99.05
&
99.28
&
&
98.20
&
98.52
&
98.67
&
98.85
&
99.03
&
\\

\rowcolor{gray!10}[0pt][0pt]
\textbf{OODA}
&
\textbf{Specialized}
&
\textbf{85.10}
&
\textbf{85.46}
&
\textbf{87.30}
&
\textbf{89.52}
&
\textbf{94.90}
&
&
\textbf{98.55}
&
\textbf{99.09}
&
\textbf{99.16}
&
\textbf{99.30}
&
\textbf{99.42}
&
&
\textbf{98.40}
&
\textbf{98.73}
&
\textbf{98.86}
&
\textbf{98.98}
&
\textbf{99.10}
&
\\

\bottomrule
\end{tabular}%
}

\caption{
ToolDial results across Qwen3 backbone sizes. Specialized OODA achieves the
highest Task Success at every scale, with larger gains on smaller backbones.
Bold indicates the best result, and underlining indicates the strongest non-OODA baseline.
}
\label{tab:tooldial_main}
\end{table*}

\subsection{Experimental Setup}

\paragraph{Data, splits, and supervision.}
We use the official ToolDial split of 11{,}111 multi-turn tool-use
sessions~\citep{DBLP:conf/iclr/ShimSLJ25}. Trajectories are split before
label construction, keeping all turns from a session in the same partition.
Supervision is constructed from the visible interaction prefix and the
current-turn gold action, while excluding future user turns, the observation
returned by the current gold action, and facts revealed only by that
observation. Each trajectory is converted into aligned training views for
Direct-LoRA, ReAct-LoRA, $\alpha$-UMi, Joint OODA, Shared OODA, and
Specialized OODA. All systems use the same tokenizer and context budget;
examples are tokenized offline, grouped by length, dynamically padded, and
never packed across turn boundaries.

\paragraph{Systems and training.}
We evaluate Qwen3-Instruct backbones from 0.6B to 14B
parameters~\citep{DBLP:journals/corr/abs-2505-09388}.
Direct-LoRA predicts the external output in one pass, ReAct-LoRA generates a
free-form thought before acting, Direct-SC@4 aggregates four Direct-LoRA
samples, and $\alpha$-UMi uses Planner, Caller, and Summarizer
roles~\citep{DBLP:conf/iclr/HuSWALWWC22,DBLP:conf/iclr/YaoZYDSN023,
DBLP:conf/iclr/0002WSLCNCZ23,DBLP:conf/emnlp/ShenLCYQCZ024}.
Joint, Shared, and Specialized OODA respectively use one-pass typed
generation, four stage-wise calls with a shared adapter, and four
stage-specific adapters.

At each scale, systems share the Qwen3-Instruct backbone, set
\texttt{enable\_thinking=false}, and use an 8192-token limit. All LoRA-tuned
systems use rank $r=32$, scaling $\alpha=64$, 0.05 dropout, and
\texttt{target\_modules=all-linear}, with the backbone, embeddings, and
\texttt{lm\_head} frozen. Optimizer family, learning-rate schedule, epoch
count, validation procedure, and output processing are shared; training uses
BF16 and FlashAttention~2 \citep{DBLP:conf/iclr/Dao24}. Per-device batch size
and gradient accumulation vary by scale to keep the global batch approximately
constant. Checkpoints and decoding settings are selected on validation data.
\paragraph{Constraint profiles.}
Each \textsc{ConstraintPolicy} generates a separate typed training set from
the same trajectories. Profiles share the data split, architecture, LoRA
configuration, optimizer, training schedule, and token-level
cross-entropy objective. At evaluation time, each model is paired with the
controller corresponding to its training profile.

\paragraph{Evaluation pipeline.}
All methods use the same output parser, schema normalizer, validator,
benchmark adapter, and metric implementation. Shared metrics are computed
from the final external action or response; OODA intermediate states are
used only for targeted diagnostics. The main text emphasizes cross-scale
patterns and representative mechanisms, while complete results
are reported in the supplement.

\paragraph{External datasets.}
For transfer experiments, models retain their ToolDial-trained adapters and
receive no additional training examples or in-context demonstrations on
FAIL-TaLMs, MTU-Bench, or BFCL
\citep{DBLP:conf/naacl/TrevinoCNNW25,DBLP:conf/iclr/WangWWLSPDZWPZG25,
DBLP:conf/icml/PatilMYJSSG25}.
All external evaluations use greedy decoding and the same
benchmark-specific conversion and scoring pipeline across methods.

\paragraph{Metrics and statistics.}
ToolDial is evaluated with Task Success, Tool Exact, and Ask--Act Accuracy.
On tool-call turns, Task Success requires \texttt{tool\_exact},
\texttt{argument\_exact}, and \texttt{schema\_valid};
\texttt{grounding\_valid} is reported separately as a typed-state metric.
For methods with typed intermediate states, we additionally report State
Slot F1 and Typed Constraint Consistency. Across all methods, we measure
Ask-vs-Act F1, premature-call rate, and stale-or-ungrounded-argument rate,
with slices by history length, missing information, state changes, active
constraints, dependency depth, and parallel-call structure. MTU-Bench uses
the metrics provided by MTU-Eval, including S-M and S-S averages.
FAIL-TaLMs reports Ask--Act Accuracy, Schema Validity, Premature Call, and
Required-Argument Coverage; typed-state methods additionally report Trace
Grounding Validity, Readiness Contract Validity, Slot Exactness, and Stale
Binding. BFCL reports its official Overall, Non-Live, Live, and Multi-turn
tracks.

\subsection{\rqhighlight{red!15}{RQ1: Does OODA Improve Tool Use?}}

Table~\ref{tab:tooldial_main} shows that Specialized OODA consistently
achieves the best Task Success across all model scales. Its advantage is
largest for smaller backbones and narrows as capacity grows, indicating that
explicit decision-state decomposition is especially useful under limited
model capacity. Tool Exact follows the same general pattern, whereas
Ask--Act Accuracy is already near saturation for structured methods. The
progression from Joint to Shared and then Specialized OODA further suggests
that both multi-pass decomposition and stage-specific adaptation contribute
to the gains.

\subsection{\rqhighlight{yellow!15}{RQ2: When Does OODA Help Most?}}

\paragraph{Capacity.}
Across the hard and out-of-distribution evaluation subsets, Specialized OODA
remains the strongest method at every reported backbone size; detailed results
appear in \appsecref{\AppHardOOD}. Absolute performance improves with scale,
but the margin over the strongest size-matched one-pass baseline steadily
narrows. Together with Table~\ref{tab:tooldial_main}, this pattern suggests
that larger backbones partially recover the state-tracking and decision
capacity supplied explicitly by OODA. These subset results characterize model
behavior; overall results are reported in Table~\ref{tab:tooldial_main}, and
slice definitions and sample counts in \appsecref{\AppSliceConstruction}.

\begin{figure}[t]
\centering
\includegraphics[width=\linewidth]{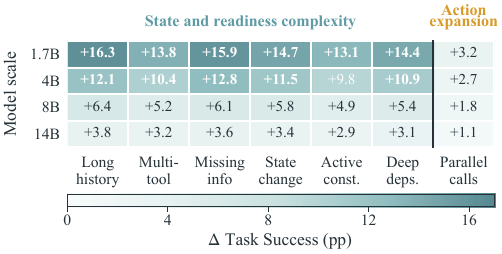}
\caption{
Task Success gains of Specialized OODA over Direct-LoRA across model scales
and ToolDial complexity subsets. Gains are largest on state-intensive tasks
and smallest on parallel calls; values are percentage-point differences.
}
\label{fig:complexity_gain}
\end{figure}

\begin{figure}[t]
\centering

\begin{minipage}[t]{\linewidth}
    \centering
    \includegraphics[width=\linewidth]{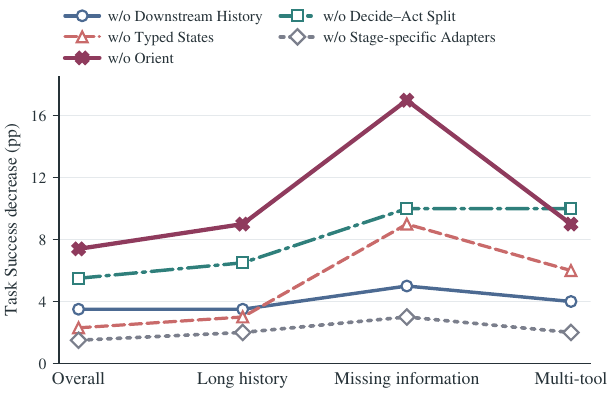}
    \par
    \smallskip
    {\small\textbf{(a) Component ablations}\par}
\end{minipage}

\medskip

\begin{minipage}[t]{\linewidth}
    \centering
    \includegraphics[width=\linewidth]
    {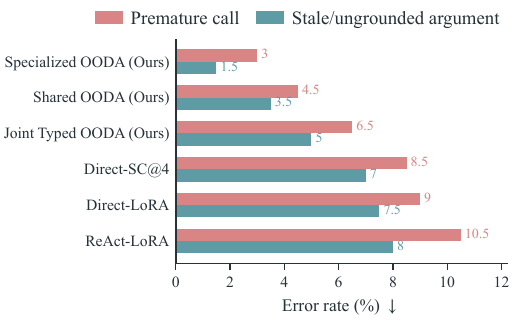}
    \par
    \smallskip
    {\small\textbf{(b) Grounding Errors}\par}
\end{minipage}

\caption{
Component ablations and execution-grounding errors. Panel (a) reports Task
Success decreases under controlled ablations, and panel (b) reports
premature-call and stale-or-ungrounded-argument rates; larger ablation drops
are worse, while lower error rates are better. Orient and the Decide--Act
separation contribute most, whereas stage-specific adapters have a smaller
effect.
}
\label{fig:mechanism_summary}
\end{figure}

\paragraph{Task structure.}
Figure~\ref{fig:complexity_gain} shows that OODA's gains concentrate on turns
requiring state reconstruction: long histories, missing required
information, state changes, active constraints, and deeper tool
dependencies. The improvement is much smaller for parallel calls, where the
dominant errors concern call expansion and cross-call argument binding rather
than dialogue-state reconstruction. OODA therefore addresses state-dependent
reasoning more effectively than parallel action realization.

\subsection{\rqhighlight{green!15}{RQ3: Which Components Matter?}}

In ablation analyses, \textbf{Full OODA} refers to
Specialized OODA without any component removed. We compare five
variants: \textbf{w/o Orient} removes the Orient stage;
\textbf{w/o Decide--Act Split} merges decision formation with action
realization; \textbf{w/o Typed States} replaces typed intermediate states
with free-form text; \textbf{w/o Downstream History} restricts downstream
stages to typed upstream states and tool schemas; and
\textbf{w/o Stage-specific Adapters} shares one adapter across all stages.

\paragraph{Ablation sensitivity.}
The mechanism set contains 1{,}250 turns from 600 dialogues.
Figure~\ref{fig:mechanism_summary}(a) shows that \textbf{w/o Orient} causes
the largest degradation, especially when required information is missing.
\textbf{w/o Decide--Act Split} is also particularly harmful on
missing-information and multi-tool turns, showing that deciding whether and
what to call should remain distinct from surface-level action generation.
\textbf{w/o Typed States} remains competitive in aggregate but degrades
sharply on these structured subsets. By contrast,
\textbf{w/o Downstream History} causes a moderate loss, while
\textbf{w/o Stage-specific Adapters} has the smallest effect. The main
benefit therefore comes from explicit staged state construction, with
stage-specific adaptation providing an additional but smaller gain.

\paragraph{Execution grounding.}
Figure~\ref{fig:mechanism_summary}(b) evaluates all methods on a fixed
500-turn sample drawn from rare-error pools. Specialized OODA achieves the
lowest premature-call and stale-or-ungrounded-argument rates. Its advantage
thus extends beyond producing syntactically valid calls: the staged typed
state also improves when a call is issued and whether its arguments are
supported by the interaction history.

\subsection{\rqhighlight{blue!15}{RQ4: What Transfers, and at What Cost?}}

\begin{figure}[t!]
\centering
\includegraphics[width=0.99\columnwidth]{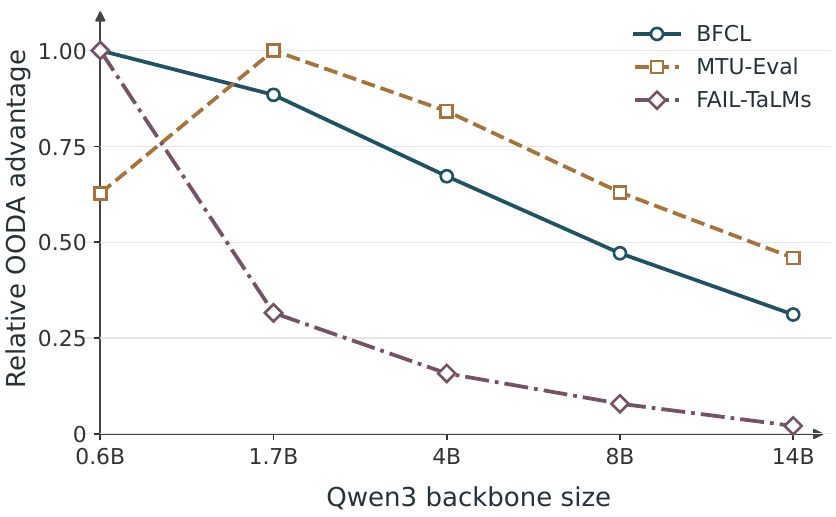}
\caption{
Normalized transfer advantage of Specialized OODA over $\alpha$-UMi across
Qwen3 backbone sizes on BFCL, MTU-Eval, and FAIL-TaLMs. Higher values indicate
larger gains. The advantage is strongest on BFCL and smaller backbones and
narrows with scale.
}
\label{fig:external_transfer_scale}
\end{figure}

We use FAIL-TaLMs to evaluate readiness and failure handling, MTU-Bench to
measure transfer across tool-use settings, and BFCL to evaluate call
realization. Figure~\ref{fig:external_transfer_scale} summarizes the
cross-scale margin of Specialized OODA over $\alpha$-UMi; complete tables
are reported in the supplement.

\paragraph{Zero-shot readiness transfer: FAIL-TaLMs.}
At 1.7B, Specialized OODA is saturated on the shared readiness
metrics and is best or tied with the strongest baselines. Its margin over
$\alpha$-UMi becomes negligible at larger scales, indicating that FAIL-TaLMs
mainly tests capabilities that sufficiently large structured models already
solve. Typed-state diagnostics likewise show near-perfect grounding,
readiness, slot accuracy, and stale-binding control. These internally
evaluated results are not official FAIL-TaLMs leaderboard scores, and the
prediction summary does not support a unified setting-level success
table. Full results appear in \appsecref{\AppFAIL}.

\paragraph{Cross-setting transfer: MTU-Bench.}
Specialized OODA improves most MTU-Bench metrics at 1.7B while tying
the remaining ones. The advantage is consistent but modest, suggesting that
the learned state decomposition transfers beyond ToolDial without dominating
every aspect of multi-turn tool use. Complete all-scale and per-metric results
appear in \appsecref{\AppMTU}.

\paragraph{Call realization: BFCL.}
BFCL shows the clearest benefit for smaller
backbones. The margin decreases with scale, indicating that explicit
state decomposition compensates when backbone capacity is limited.
However, the smaller gain on the Multi-turn track is consistent with
parallel-call errors observed on ToolDial. Full results appear in
\appsecref{\AppBFCL}.

\paragraph{Efficiency trade-off.}
At 1.7B, Specialized OODA uses four sequential calls with
2.36$\times$ normalized latency relative to Direct-LoRA. Its
accuracy--latency trade-off is most favorable on turns involving missing
information, state changes, active constraints, or tool dependencies, while
Direct-LoRA remains suitable for simple latency-sensitive calls. Direct-SC@4
uses the same number of model calls but remains below the OODA variants,
indicating that the gain comes from structured computation rather than
repeated sampling. Parallel-call realization remains a shared limitation of
structured methods. Detailed cost measurements and the accuracy--latency
analysis appear in \appsecref{\AppEfficiency}.

\subsection{\rqhighlight{violet!15}{
  \RQTypedState: Can Typed States Support Action?
}}

\paragraph{Typed-state sufficiency.}
\textbf{w/o Downstream History} retains most of Full OODA's performance
without dialogue-history access, but degrades on long-history,
missing-information, and multi-tool turns, suggesting that typed states
preserve most but not all task-relevant context. \textbf{w/o Typed States}
is less robust, while \textbf{w/o Orient} causes the largest losses in state
reconstruction and constraint consistency despite near-ceiling Ask-vs-Act
accuracy. Thus, free-form states' high Schema Validity does not imply
equivalent state quality (\appsecref{\AppMechanism}).

\paragraph{State--action alignment.}
On turns where all variants recover the correct action-relevant state and
Orient mode, the State--Action Contradiction Rate (SACR; lower is better)
falls from 7.8\% for Joint and 5.6\% for Shared OODA to 3.9\% for
Specialized OODA, but rises to 10.7\% without the Decide--Act split.
This 6.8-point gap indicates that separating action selection from realization
reduces residual response-mode, superseded-value, and unsupported-binding
errors. A profile comparison varying only readiness, active-state, and
argument-grounding targets preserves perfect schema validity while the
constraint-enriched profile removes stale bindings and restores grounding.
Gradient alignment also increases during training, especially for Act-related
pairs. Further analyses appear in the supplement sections
\emph{State--Action Competition},
\emph{Constraint-Policy Alignment Across Training and Inference}, and
\emph{Gradient Analysis Protocol}.

\paragraph{Interpretation and boundary.}
Typed states support action when readiness, active values, and admissible
sources are represented consistently. The main limitation is action expansion:
gains are smaller on BFCL Multi-turn, while ToolDial parallel calls still
expose call-expansion and cross-call binding errors.

\subsection{\rqhighlight{orange!15}{RQ6: How Does OODA Ground Arguments?}}
\label{sec:case_study_identifier}

\begin{figure}[t]
\centering
\includegraphics[width=\columnwidth]{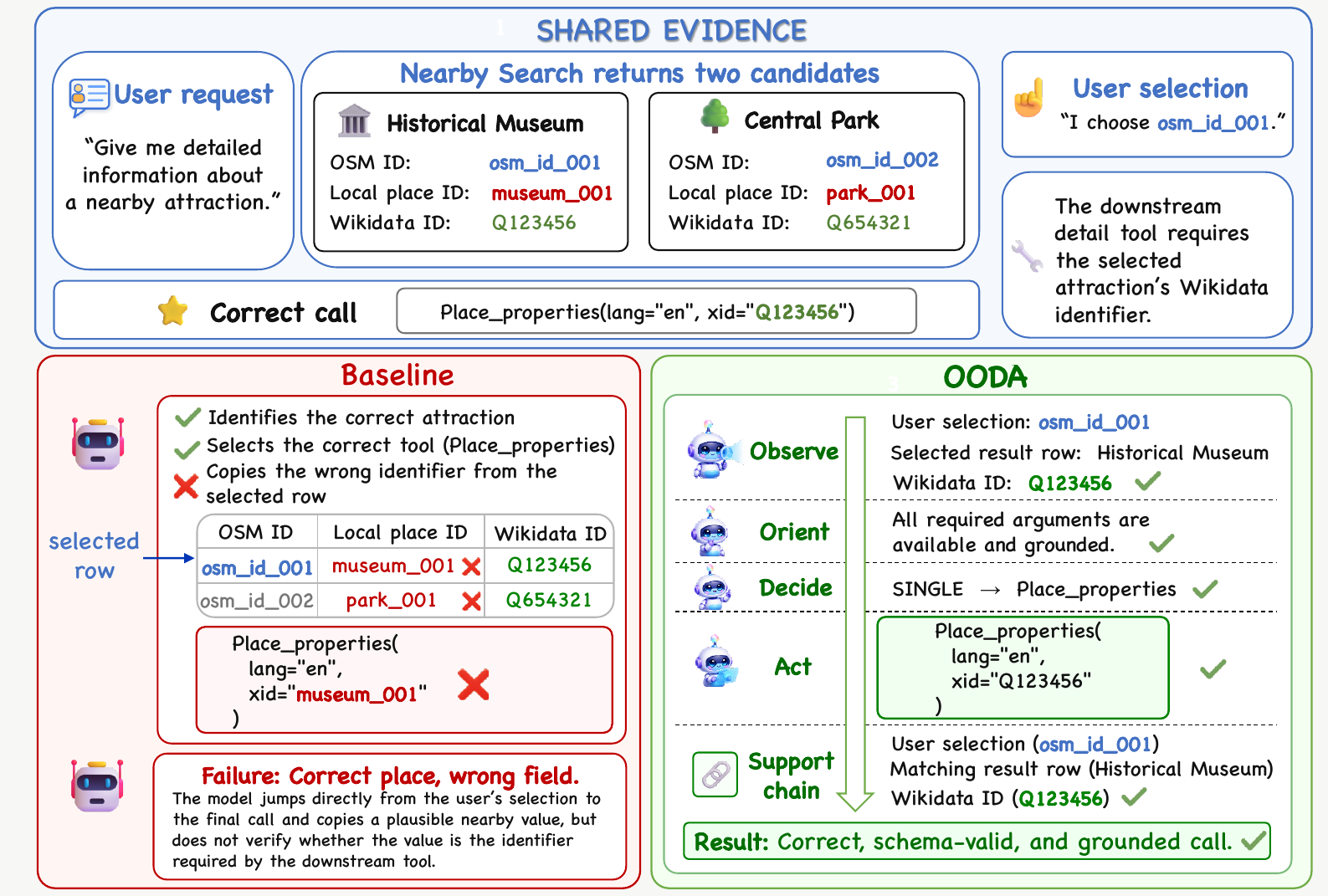}
\caption{
Identifier grounding. Baselines copy a local ID, whereas
OODA traces the selection to the Wikidata ID.
}
\label{fig:case_wrong_identifier}
\end{figure}

Figure~\ref{fig:case_wrong_identifier} illustrates a ToolDial
identifier-grounding failure. The user selects \texttt{osm\_id\_001}, but
\texttt{Place\_properties} requires a Wikidata \texttt{xid}. Direct-LoRA and
ReAct-LoRA choose the correct attraction and tool yet copy
\texttt{museum\_001}. OODA resolves the selected row in \textsc{Observe},
extracts \texttt{Q123456}, and produces the grounded call, preventing
cross-field identifier substitution.


\section{Conclusion}

We introduced OODA-Tool, a typed closed-loop policy that mitigates
state--action competition by separating task-state preservation from action
realization through controller-checked Observe, Orient, Decide, and Act stages.
Across scales, OODA-Tool improves multi-turn tool use for smaller models
and state-intensive tasks involving incomplete information, changing
constraints, and sequential dependencies. Ablations show that typed state
construction and readiness--action separation drive gains, while transfer
results suggest that these interfaces remain useful beyond ToolDial despite
sequential inference cost. Smaller gains on parallel calls identify
action expansion and cross-call binding as key challenges.

\clearpage
\bibliography{reference}

\clearpage
\includepdf[pages=-,pagecommand={},fitpaper=true]{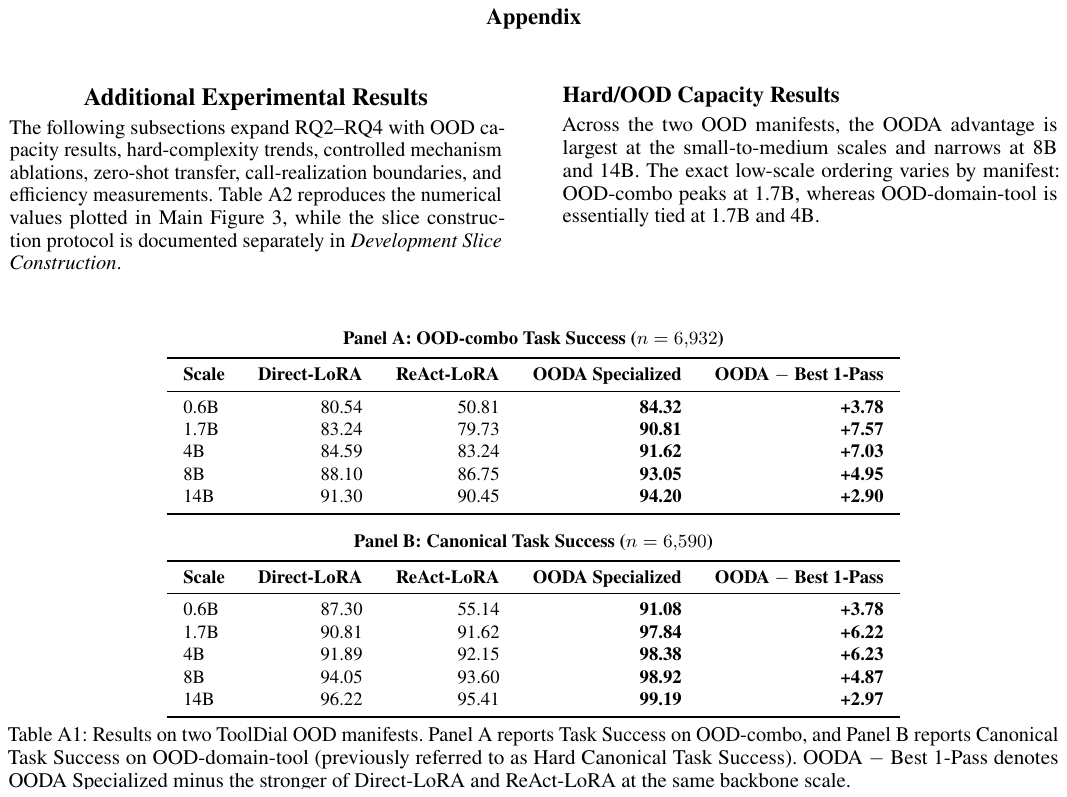}

\end{document}